\documentclass[10pt,twocolumn,letterpaper]{article}

\usepackage{cvpr}

\definecolor{cvprblue}{rgb}{0.21,0.49,0.74}
\usepackage[pagebackref,breaklinks,colorlinks,allcolors=cvprblue]{hyperref}
\usepackage{booktabs}
\usepackage{multirow}
\usepackage{siunitx}
\usepackage{adjustbox}
\usepackage{algorithm}
\usepackage{caption}
\usepackage{amsmath}

\usepackage{graphicx}
\usepackage{caption}
\usepackage{multirow}
\usepackage{titletoc}
\usepackage{enumitem} 
\usepackage{colortbl}
\usepackage{booktabs}
\usepackage{animate}

\usepackage{makecell}

\usepackage{color}

\usepackage{colortbl}
\usepackage{wrapfig}
\usepackage{csquotes}
\usepackage{amsmath}
\usepackage{algpseudocode}
\usepackage{appendix}
\usepackage{pifont}
\usepackage{makecell}

\newcommand\blfootnote[1]{%
  \begingroup
  \renewcommand\thefootnote{}\footnote{#1}%
  \addtocounter{footnote}{-1}%
  \endgroup
}

\definecolor{Red}{RGB}{192, 0, 0}
\definecolor{Blue}{RGB}{12, 114, 186} 

\title{InstanceAnimator: Multi-Instance Sketch Video Colorization}

\author{
Yinhan Zhang\textsuperscript{1*} \quad 
Yue Ma\textsuperscript{2*} \quad 
Bingyuan Wang\textsuperscript{1} \quad 
Kunyu Feng\textsuperscript{1} \\
Yeying Jin\textsuperscript{3} \quad 
Qifeng Chen\textsuperscript{2} \quad 
Anyi Rao\textsuperscript{2} \quad 
Zeyu Wang\textsuperscript{1,2\dag}
\\
\\
\textsuperscript{1} HKUST(GZ),
\textsuperscript{2} HKUST,
\textsuperscript{3} NUS
\\
\\
Project Page: \url{https://yinhan-zhang.github.io/animator}
}
\begin{document}

\maketitle
\blfootnote{* Equal contribution.}
\blfootnote{\textsuperscript{\dag} Corresponding author.}

\begin{abstract}
We propose \textit{InstanceAnimator}, a novel Diffusion Transformer framework for multi-instance sketch video colorization. Existing methods suffer from three core limitations: inflexible user control due to heavy reliance on single reference frames, poor instance controllability leading to misalignment in multi-character scenarios, and degraded detail fidelity in fine-grained regions. To address these challenges, we introduce three corresponding innovations. First, a Canvas Guidance Condition eliminates workflow fragmentation by allowing free placement of reference elements and background, enabling unprecedented user flexibility. Second, an Instance Matching Mechanism resolves misalignment by integrating instance features with the sketches, ensuring precise control over multiple characters. Third, an Adaptive Decoupled Control Module enhances detail fidelity by injecting semantic features from characters, backgrounds, and text conditions into the diffusion process. Extensive experiments demonstrate that \textit{InstanceAnimator} achieves superior multi-instance colorization with enhanced user control, high visual quality, and strong instance consistency.
\end{abstract}    

\section{Introduction}
\begin{figure*}[t]
    \centering
    \includegraphics[width=\linewidth]{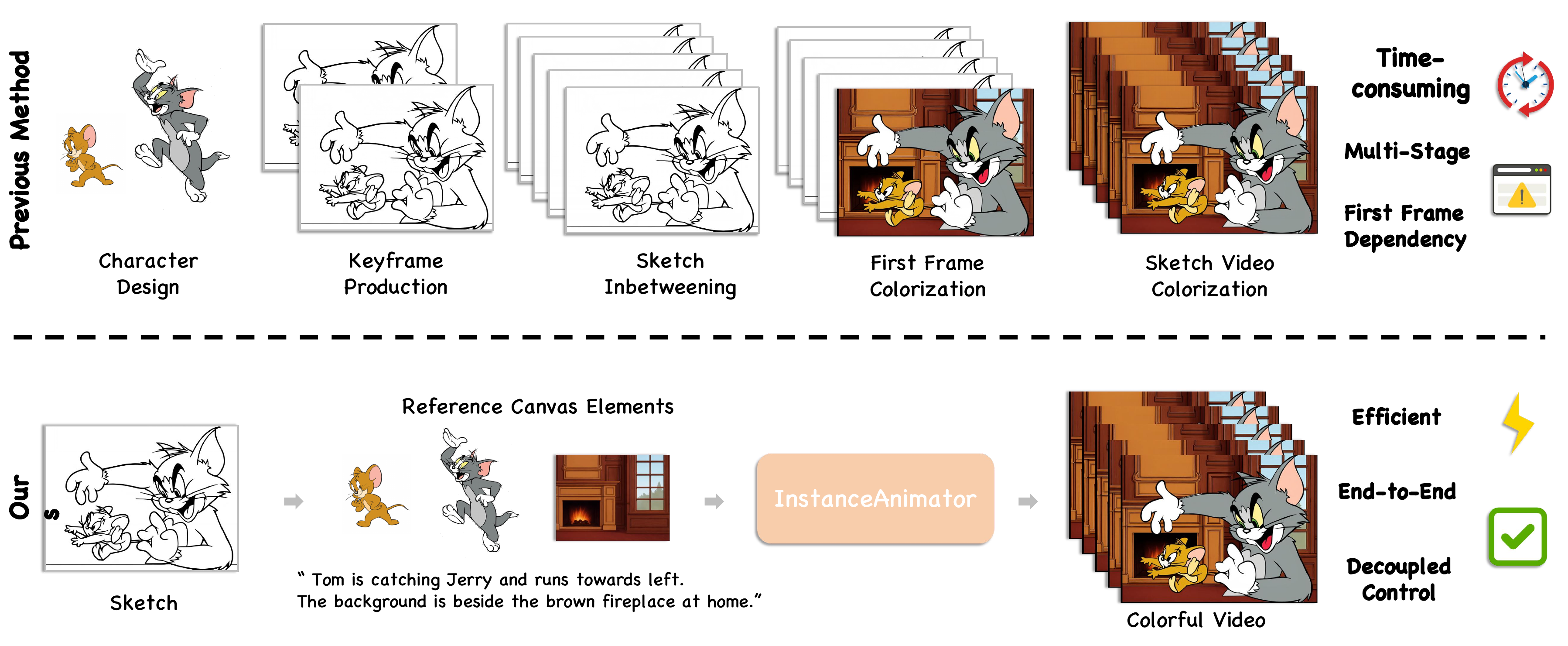}
    \caption{\textbf{Motivation}. Unlike traditional methods that require multi-stage, time-consuming, and first-frame-dependent colorization, \textit{InstanceAnimator} directly colorizes sketch sequences into videos after background and character design. Our method no longer relies on a single reference keyframe but adjusts end-to-end instance-aware colorization, providing decoupled control to achieve higher user flexibility and significantly reducing time consumption and labor.}
    \label{fig: introduction}
\end{figure*}

Animation plays a crucial role in visual content creation and is widely applied in education, cinema, entertainment, and digital media creation to convey coherent, dynamic narratives. As shown in Figure~\ref{fig: introduction}, the colorization of animation is commonly built on a meticulous and iterative process, including character design, keyframe production, frame inbetweening, and per-element colorization. Although traditional multi-stage colorization methods have allowed fine-grained control by these stages~\citep{isola2017image,li2021deep,shen2022enhanced, cen2025layert2v, longanimation}, they still demand intensive artistic labor, especially for multi-character scenes. This multi-stage approach also requires sketching and colorizing hundreds of frames for only a few seconds of animation, imposing significant time and labor costs.

Recent advances in the diffusion models~\citep{rombach2022high, peebles2023scalable} have significantly enhanced the quality and flexibility of content generation. Based on these pretrained models, researchers proposed a new approach to automate animation colorization using reference images to guide color propagation~\citep{tang2025generative, tooncomposer, meng2024anidoc, yang2025layeranimatelayerlevelcontrolanimation}. However, the core issue with these methods is their heavy dependency on a single reference frame as an exemplar, which prevents the dynamic customization capabilities that professional animators require. This rigid constraint also hinders artists from performing creative tasks, such as independently modifying a character's costume color, adjusting background lighting to fit plots and scenarios, or experimenting with alternative texture schemes. In previous methods, such changes require reprocessing the entire scene, which is inconsistent with the basic practices in production and common requirements from animators---\textbf{the iterative refinement of discrete visual elements}. Artists are thus commonly forced to sacrifice either expressiveness or efficiency in their creative process. We summarize the limitations of existing video line art colorization methods as follows:
\begin{enumerate}[label=(\arabic*), nosep, leftmargin=*]
    \item \textit{\textbf{First frame dependency}}: Current methods' reliance on a single initial reference frame causes three major limitations. First, it restricts creative flexibility by preventing independent modification of visual elements (e.g., character costumes, lighting). Second, it fails to handle large character motions (\textit{e.g.}, running, jumping) or camera movements (e.g., panning shots), where the fixed reference becomes inadequate for drastically different poses or viewpoints. Third, it limits results to short sequences, reducing practical applicability in real production settings.

    \item \textit{\textbf{Mismatch between instances and sketches}}: Existing methods, particularly Diffusion Transformer (DiT)-based frameworks, cannot establish reliable correspondence between reference instances (\textit{e.g.}, character design sheets) and sketch sequences. Unlike UNet-based models that leverage pixel-level feature matching to enforce spatial alignment, DiTs lack mechanisms for fine-grained visual correspondence, often resulting in unstable or inconsistent appearance.

    \item \textit{\textbf{Detail degradation}}: Existing methods often fail to preserve fine-grained visual details during propagation. Elements such as background textures, small props, clothing patterns, and subtle facial expressions are often omitted or blurred, reducing the usability of these methods in professional workflows where such details are crucial for coherence, fidelity, and expressiveness.

\end{enumerate}

To address these limitations, we propose \textit{InstanceAnimator}, a novel method for multi-instance sketch video colorization. As shown in the Figure~\ref{fig: introduction}, our approach is designed to address the rigidity of prior exemplar-driven pipelines, enabling flexible scene construction, robust instance consistency, and fine-grained detail preservation.
Unlike traditional pipelines that rely on a fixed first frame as reference, we propose a \textbf{Canvas Guidance Condition} that allows users to freely place pre-designed reference elements (\textit{e.g.}, characters, background components) onto a blank canvas. This canvas acts as a spatio-temporal anchor, enabling more natural propagation of appearance across sequences, particularly under large-scale character motions or camera movements. 
To tackle the mismatch between the reference instances and sketch sequences, we introduce an \textbf{Instance-Aware Attention}. Specifically, we concatenate instances with the sketch, compelling the attention module to learn their visual relationships and form a stronger conditional prior. Furthermore, we inject instances into the noise channels to enhance controllability during inference, allowing text guidance to better modulate the generated content. Together, these designs ensure consistent appearance transfer while preserving the inherent controllability of diffusion models.
To mitigate the degradation of fine-grained details, we propose an \textbf{Adaptive Decoupled Control Module} that selectively injects the semantic features of characters (\textit{e.g.}, clothing patterns, subtle facial expressions) and backgrounds (\textit{e.g.}, props, textures) into the denoising process. This module strengthens fidelity at a granular level, enabling the model to preserve intricate details that are often lost in previous pipelines.

With these designs, our framework can effectively generate high-quality results with diverse instance controls. Overall, our contributions can be summarized as follows:
\begin{itemize}[nosep, leftmargin=*]
    \item We propose \textit{InstanceAnimator}, the first multi-instance sketch video colorization framework, to address complex multi-reference animation tasks that are ubiquitous in sketch colorization scenarios.
    
    \item  We introduce \textbf{Canvas Guidance} to enhance user creativity that avoids the first frame dependence, and \textbf{Instance-Aware Attention} to strengthen instance consistency by establishing attention correspondence between references and sketches. We also design an \textbf{Adaptive Decoupled Control Module} that enables fine-grained control over text, background, and instances.

    \item To advance the field of full decoupled control for animation workflow, we release the \textbf{OpenAnimate Dataset}, which comprises over 42K high-quality video clips along with their associated instances, background, reference frame, and detailed textual descriptions.
    
\end{itemize}

\section{Related Work}

\subsection{Video Diffusion Model}
In recent years, video diffusion models have evolved rapidly to tackle the complexity of spatiotemporal content generation, with architectural choices playing a crucial role in their performance~\citep{ho2020denoising, blattmann2023stable}. Early advances focused on UNet-based architectures~\citep{blattmann2023align, ma2024followpose, ma2025followcreation,  ye2023ip, zhang2023adding, xing2024make, zhu2024champ,hu2024animate,ma2024follow, ye2023ip}, which extended 2D image UNets to 3D by incorporating temporal dimensions into convolutional layers. These models leveraged spatio-temporal convolutions and frame-wise attention mechanisms to capture motion dynamics, enabling the generation of short video clips with basic temporal consistency. However, they faced challenges in scaling to longer sequences due to increased computational overhead and difficulty in modeling long-range temporal dependencies.
More recently, Diffusion Transformer (DiT)-based approaches have emerged as a powerful alternative~\citep{wang2025wan,wang2025cinemaster,wang2025multishotmaster,kong2024hunyuanvideo,yang2024cogvideox}. By replacing convolutional layers with transformer blocks, DiT-based video models excel at modeling global spatio-temporal relationships through self-attention, allowing them to handle longer sequences and more complex motions. These models typically decompose videos into spatial patches across frames, treating temporal dependencies as part of the attention context, which enhances their ability to preserve coherence across dynamic scenes. Despite their strengths, both UNet and DiT-based video diffusion models primarily focus on general video generation~\citep{uni3c, ma2026fastvmt, ma2025followyourmotion, ma2025controllable, ma2025followfaster, ma2024followyouremoji, ma2025followyourclick, long2025follow, shen2025follow, ma2022visual, feng2025dit4edit, hu2024animate, he2024cameractrl, wang2024motionctrl} and lack specialized mechanisms for domain-specific tasks, such as sketch video colorization. 
Such tasks demand strict adherence to the structural integrity of input sketches and instance-level color consistency. While recent controllable video generation models can synthesize videos guided by Pose and Canny edge features, instance-level control capabilities remain notably lacking.
This critical research gap underscores the imperative for video line art colorization tasks.

\begin{figure*}[t]
    \centering
    \includegraphics[width=\linewidth]{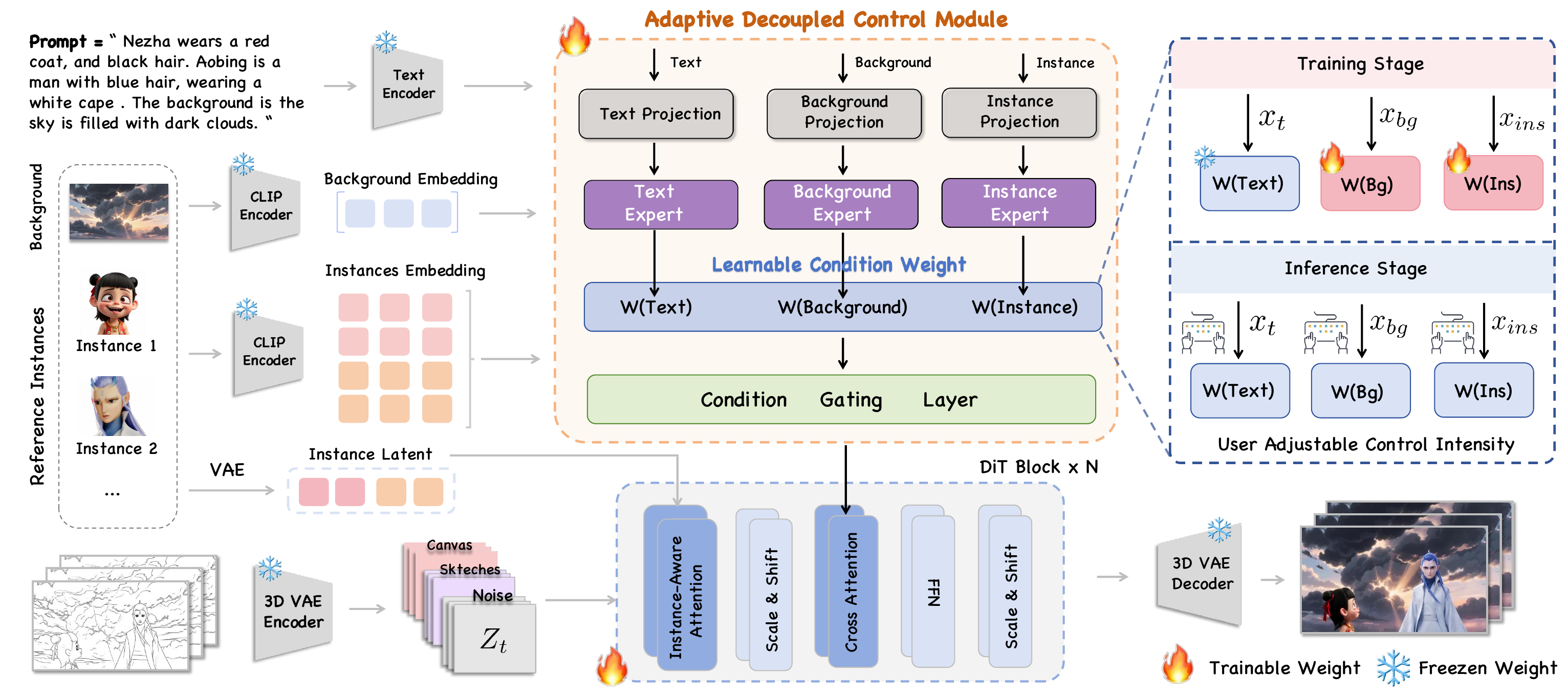}
    \caption{\textbf{Overview of InstanceAnimator.} We first apply instance-aware attention with instance latent features and noise features to establish a correspondence between the line drawing and the reference instances, as well as to maintain the character feature. Concurrently, instances, background, and text descriptions are fed into the Adaptive Decoupled Control Module independently, which dynamically injects condition information into DiT blocks through three condition-specific expert modules. At the inference stage, users can adjust the conditional weights and freely change the reference instances and image background to enhance controllability and creative flexibility.
    }
    \label{fig:framwork}
\end{figure*}

\subsection{Reference-based Line Art Colorization}
Reference-based colorization methods~\citep{zhuang2024colorflow,huang2024lvcd,xing2024tooncrafter,jiang2024exploring,tooncomposer, meng2024anidoc, followyourcolor, longanimation} aim to transfer colors from a reference image to a target sketch, bridging the gap between artistic line work and colored rendering. Traditional approaches~\citep{yan2025enhancingreferencebasedsketchcolorization, lee2020reference, zou2019language, zang2024magic3dsketchcreatecolorful3d} relied on pixel-level matching or style transfer techniques, which struggled with complex sketches and diverse reference styles. 
With the development of diffusion models, recent methods~\citep{loshchilov2017decoupled, phantom, magref, hu2024animate, shen2022enhanced, li2021deep, zhang2025animecolorreferencebasedanimationcolorization, xie2025physanimatorphysicsguidedgenerativecartoon} have adopted reference-guided generation, using cross-attention mechanisms to align sketch features with reference color information. However, these models typically depend on a single reference frame to propagate colors across sequences, which works adequately for short, static sketches but encounters significant challenges in dynamic scenarios. Specifically, when characters or backgrounds undergo substantial movements, these methods often suffer from color bleeding and inconsistent instance coloring.
Moreover, while existing methods can handle single-character sketches well, they lack robust mechanisms to manage multi-instance scenarios where multiple characters or objects require distinct, consistent color schemes across frames. Current efforts for temporal coherence remain limited by fixed sequence length and difficulties in maintaining precise alignment between evolving sketch sequences and reference designs, highlighting the need for tailored multi-instance colorization frameworks.

\section{Method}

\subsection{Problem Definition}
We formalize the conditional sketch-text-image guided video colorization task as follows: given a sketch sequence $S = \{s_1, ..., s_t\}$, a text description $T$, multiple reference instances $I = \{I_1, ..., I_N\}$, and an image background $B$, our framework aims to generate colorful video frames $V_{\text{gen}} = \{v_1, ..., v_t\}$. 
The training objective is formulated as follows:
\begin{equation}
\mathcal{L} = \mathbb{E}_{t, x_0, \epsilon} \left[ \| \epsilon - \epsilon_\theta(x_t, t, S, T, I, B) \|^2 \right],
\end{equation}
where $x_0$ denotes the ground-truth video, $\epsilon \sim \mathcal{N}(0, I)$ is the Gaussian noise, and $x_t$ is the noisy latent at timestep $t$. The denoising network $\epsilon_\theta$ is optimized to predict the noise conditioned on the sketch sequence $S$, text description $T$, reference instances $I$, and image background $B$.

To address key limitations in existing methods, we propose a novel multi-instance video colorization framework. Specifically, Sec.~\ref{canvas detail} establishes spatio-temporal consistency through a unified canvas that integrates instances and background, Sec.~\ref {instance detail} enhances instance-sketch alignment via latent feature fusion, and Sec.~\ref{control detail} preserves fine-grained details through decoupled control injection.

\subsection{Canvas Guidance}
\label{canvas detail}
\begin{figure}[h]
    \centering
    \includegraphics[width=\linewidth]{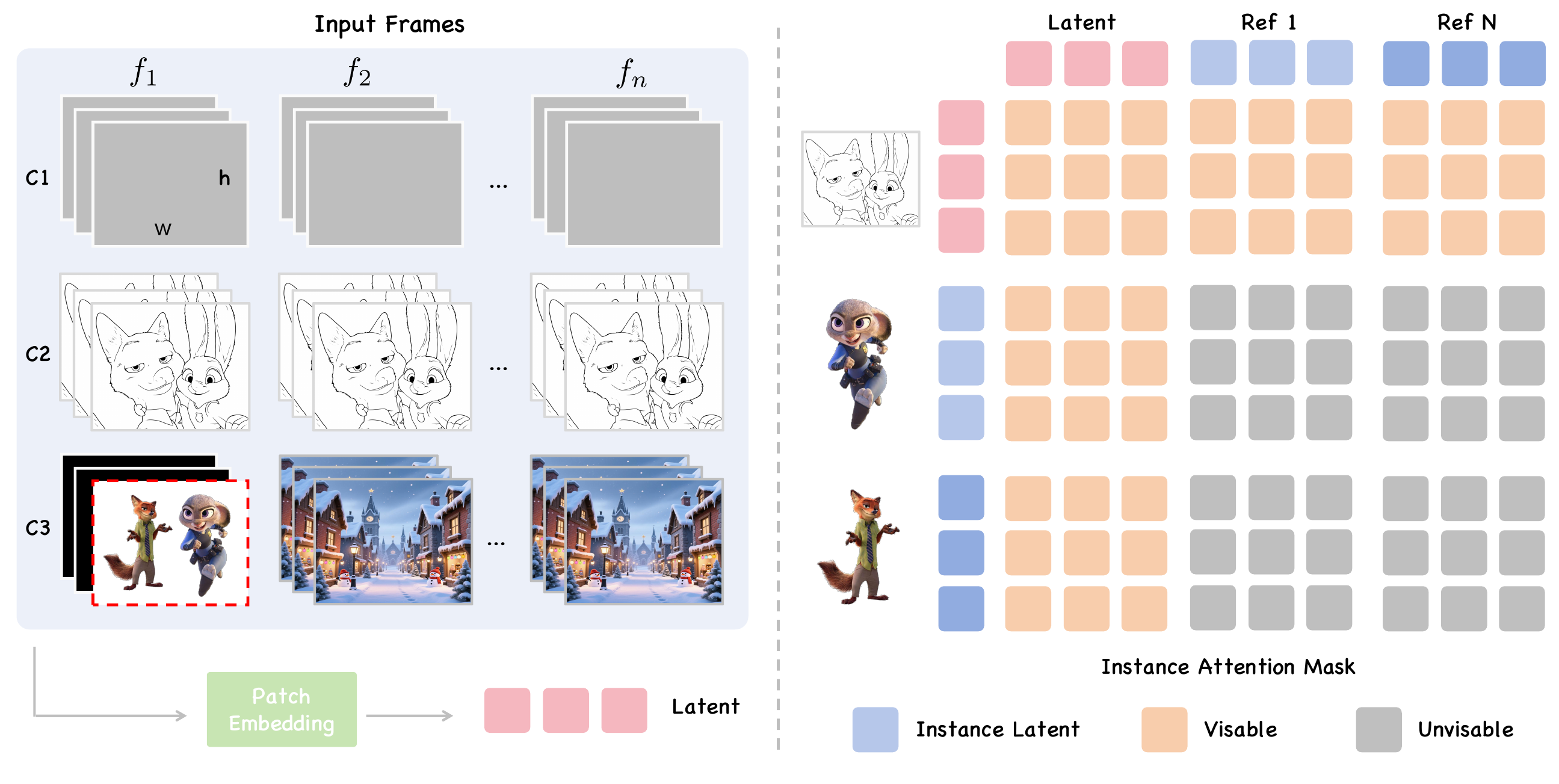}
    \caption{\textbf{(Left): Condition Fusion.} The sketch, canvas, and background conditions are temporally aligned and concatenated along the channel dimension. \textbf{(Right): Instance-Aware Attention Mask.}  This design enables the model to capture the correspondence between reference instances and sketches while avoiding a sharp increase in computational complexity.}
    \label{fig: detail}
\end{figure}

In existing methods, the reliance on the first frame as the sole reference severely constrains both character and background composition, preventing users from freely customizing these elements according to sketch inputs, thus limiting both scalability and creative flexibility.
To address this issue, we propose the Canvas Guidance to enhance the spatial consistency of instances and preserve background information. As shown in Figure~\ref{fig: detail}, the approach begins by initializing a blank canvas onto which users can place multiple pre-designed instance elements (\textit{e.g.}, characters, objects) to meet their creative requirements. For background control, the image background is used in subsequent frames as a visual reference. If there is no visual reference as input, it will fill with blank frames. This method of training can effectively promote the fusion of instance and background, and in some scenarios, it can demonstrate the coloring effect of dynamic background rendering.

This unified canvas functions as a spatio-temporal anchor, enabling the model to establish consistent temporal correlations between instances and the evolving target sequence across frames, thereby mitigating the temporal misalignment issues inherent in separate canvas encoding approaches.
Let $C \in \mathbb{R}^{H \times W \times 3}$ denote the blank canvas initialized with zeros, where $H$ and $W$ match the spatial dimensions of the target sequence. Users place $N$ instance elements $\{I_1, I_2, ..., I_N\}$ onto $C$, resulting in a composite reference canvas:  
\begin{equation}
    C_{ref} = \text{Compose}(C, \{I_1, I_2, ..., I_N\}),
\end{equation}  
where $\text{Compose}(\cdot)$ represents the placement operation that preserves the spatial positions and relationships of the instances. To integrate $C_{ref}$ into the diffusion process, we first encode it using a VAE encoder $E_{\text{VAE}}$ to obtain latent features in the VAE’s bottleneck space: $Z_{canvas} = E_{\text{VAE}}(C_{ref}, Background) \in \mathbb{R}^{H' \times W' \times D}$, where $D$ is the latent dimension and $H', W'$ corresponds to the spatial dimensions downsampled of the VAE output. 
This latent space integration ensures that the reference canvas and target sequence frames operate in a consistent feature domain.

\subsection{Instance-Aware Attention}
\label{instance detail}
Prior works primarily focus on animation colorization with first-frame reference guidance, and thus, point-based matching mechanisms between reference frames and sketches have been widely adopted in the U-Net backbone, yet such precise matching paradigms fail to generalize to DiT-based backbones and multi-reference input settings.
To address the critical need for correspondence between sketch sequences and reference instances in the DiT backbone and to underpin consistent visual alignment in dynamic scenarios, our \textit{Instance-Aware Attention} enables sketch patches to attend to instance-specific features while empowering instance tokens to aggregate spatial-contextual cues. This coordinated design, paired with dynamic adaptation capabilities, ensures reliable performance across varying multi-instance scenarios. Notably, instance tokens do not participate in the calculation of losses.

As shown in Figure~\ref{fig: detail}, we first encode reference instances to capture their distinct visual attributes. For N reference instances $\{I_1, I_2, ..., I_N\}$, we use the VAE encoder $E_{\text{VAE}}$ to generate instance-specific latent features, which act as tokens for aggregating spatial-contextual cues:  
\begin{equation}
    Z_{\text{inst}}^i = E_{\text{VAE}}(I_i) \in \mathbb{R}^{H' \times W' \times D}, \quad \forall i \in \{1, 2, ..., N\},
\end{equation}
where $D$ is the latent dimension, and  $H'$, $W'$ align with the spatial dimensions of sketch and canvas features.  Next, we fuse sketch and canvas components into a unified joint feature to enable cross-modal interaction. For a given timestep t, let $Z_{\text{sketch}}^t \in \mathbb{R}^{H' \times W' \times D}$ denote the sketch latent, and $Z_{\text{canvas}} \in \mathbb{R}^{H' \times W' \times D}$ denote the canvas latent (providing global semantic priors from Canvas Guidance). As shown in Figure~\ref{fig: detail}, the joint feature is defined as:  
\begin{equation}
\underbrace{Z_{\mathrm{joint}}^{(t)}}_{\mathbb{R}^{D_j}} = \left[ 
  \underbrace{Z_{\mathrm{noise}}^{(t)}}_{\mathbb{R}^{D_n}} 
  \parallel 
  \underbrace{Z_{\mathrm{sketch}}^{(t)}}_{\mathbb{R}^{D_s}} 
  \parallel 
  \underbrace{Z_{\mathrm{canvas}}}_{\mathbb{R}^{D_c}} 
\right],
\end{equation} 
where $Z_{\text{noise}}^t \in \mathbb{R}^{H' \times W' \times D}$ is the noise latent, $D_j$ is the total dimension of the joint feature, and $\parallel$ denotes channel-wise concatenation.  
This $Z_{\text{joint}}^t$ serves as input to DiT along with the instance features. To prevent interference among different reference instances, we employ an instance-grouped attention scheme, where each instance feature independently interacts with the joint feature without attending to other instances. Formally, for the i-th instance feature $Z_{\text{inst}}^i$, the attention is computed as:

\begin{equation}
    \text{InstanceAttn}{(i)} = \text{softmax}\left( \frac{Q_{\text{joint}} \, K_{\text{inst}}^{(i) \top}}{\sqrt{D_k}} \right) V_{\text{inst}}^{(i)},
\end{equation}
where $Q_{\text{joint}}$ is derived from $Z_{\text{joint}}^t$, and $K_{\text{inst}}^{(i)}$, $V_{\text{inst}}^{(i)}$ are obtained from $Z_{\text{inst}}^i$. This process is performed independently for each i, ensuring that the i-th instance only aggregates information from the sketch, canvas, and noise latents, but not from other instance features.  
This design strengthens correspondence by encoding semantic and spatial relationships between sketches, individual instances, and the global context, while preserving instance-specific visual attributes.

\begin{figure}[htbp]
    \centering
    \begin{minipage}{0.48\textwidth}
        \centering
        \includegraphics[width=\linewidth]{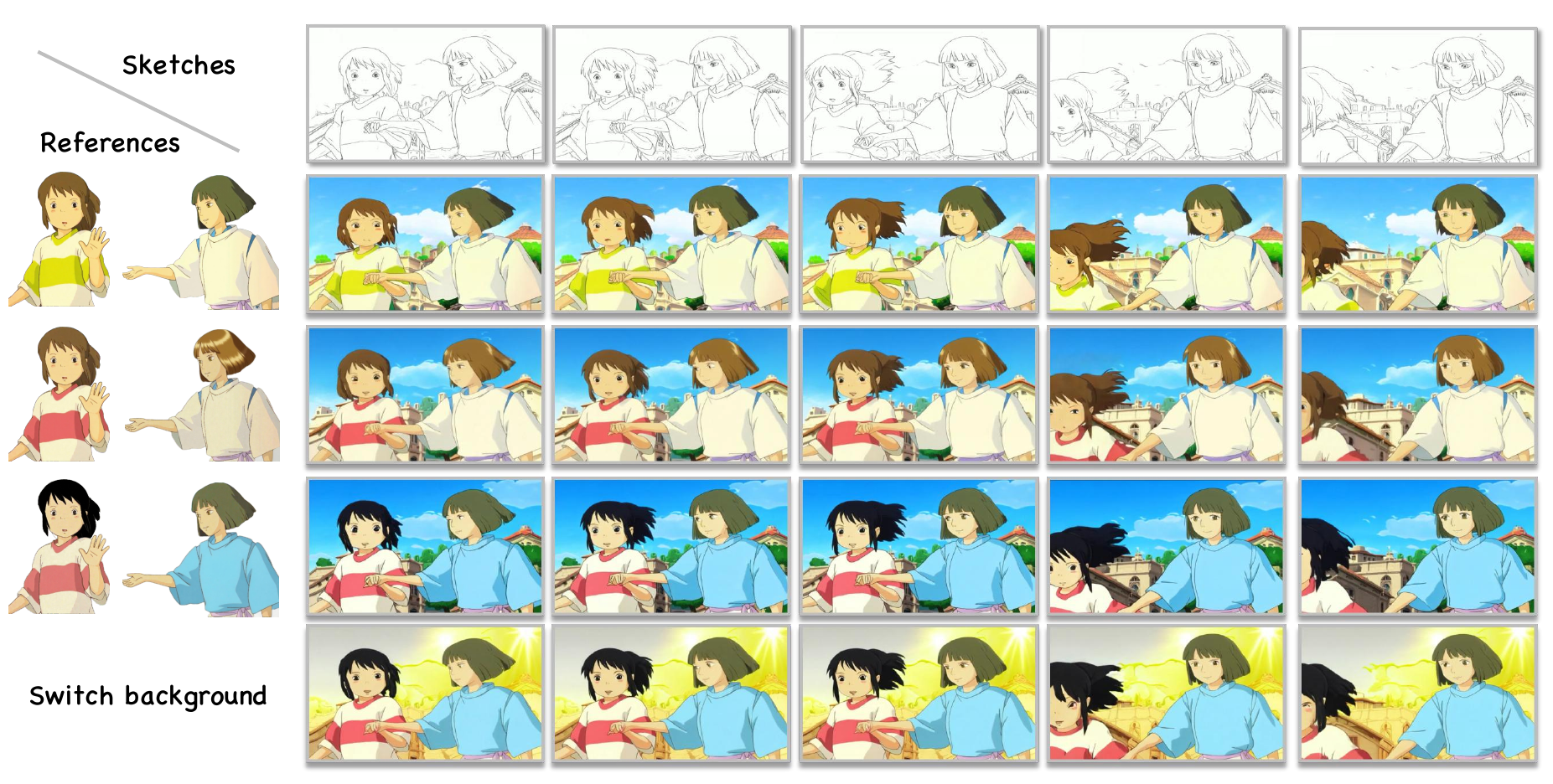}
    \end{minipage}
    \hfill
    \begin{minipage}{0.48\textwidth}
        \centering
        \includegraphics[width=\linewidth]{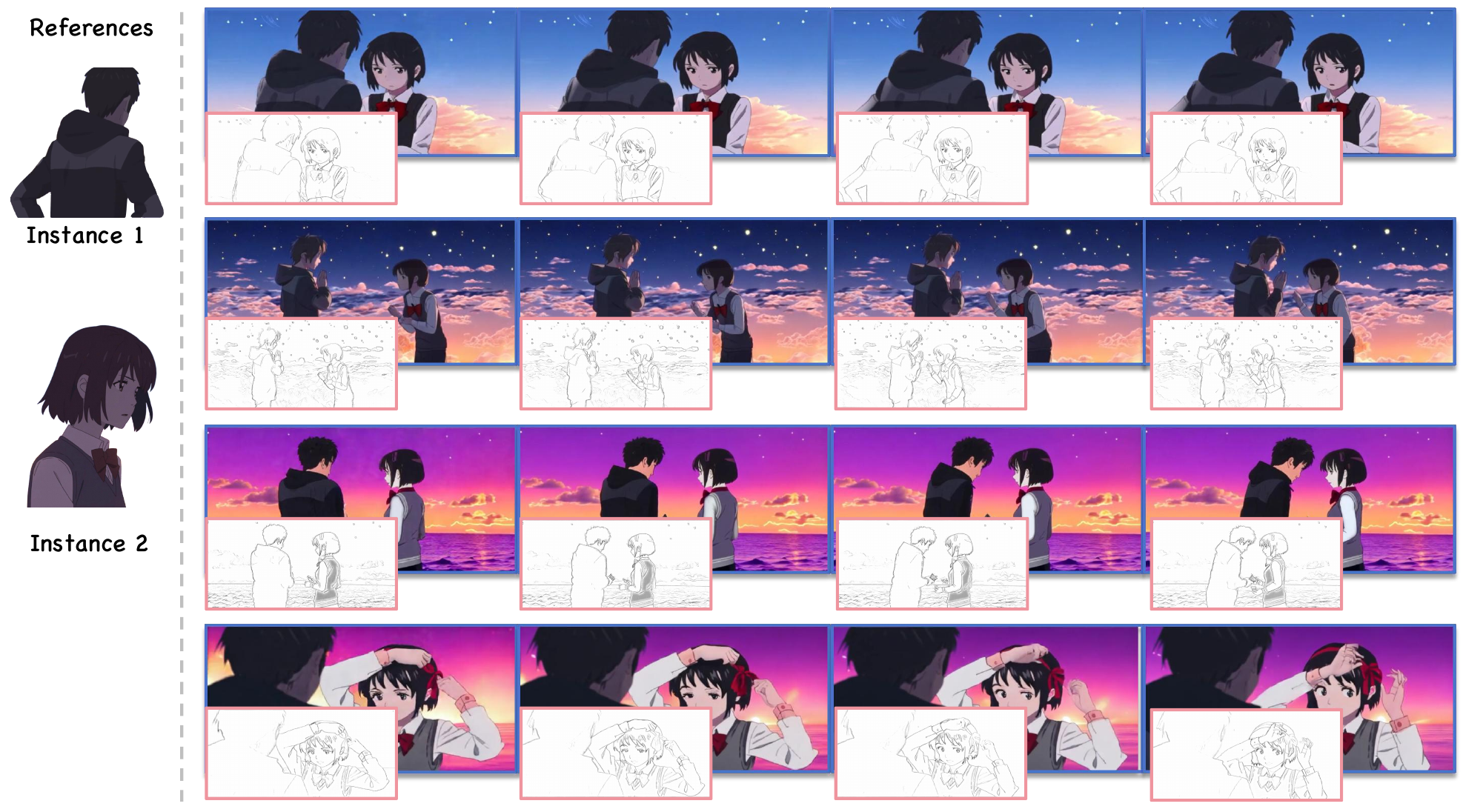}
    \end{minipage}
    \caption{\textbf{Instance Control Ability.} \textbf{(Up)} Given the same sketch and different reference instances, \textit{InstanceAnimator} generates a variety of colorful videos. \textbf{(Down)} Using the same designed characters, our framework colorizes different sketches with consistent colors and user-customized backgrounds.}
    \label{fig: control}
\end{figure}

As shown in Figure~\ref{fig: control}~(Up), a core advantage of this design is its instance adaptation. Our framework inherently supports arbitrary $N$ (adding, swapping, or replacing instances) without retraining. By treating instance tokens as modular conditioning signals, the self-attention module automatically adjusts to new instances, ensuring robust generalization across diverse instances. 
Notably, instance tokens and canvas features act as persistent conditioning cues. They do not interfere with core diffusion computations and do not require extra trainable parameters, further enhancing stability and adaptability.

\begin{figure}[htbp]
\centering
\includegraphics[width=\linewidth, keepaspectratio]{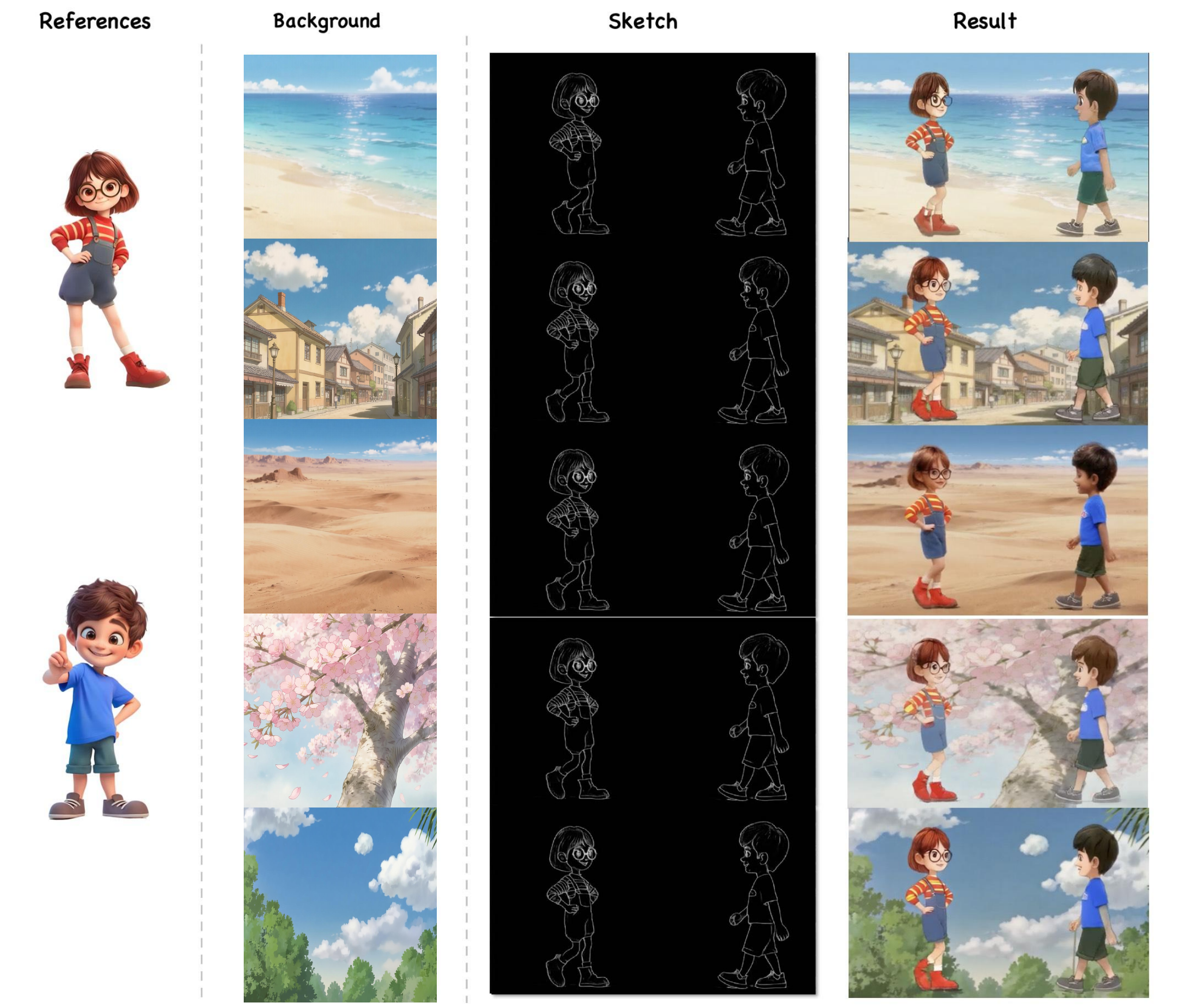}
\caption{\textbf{Visual Background Control}. Our method supports customized visual background control in combination with reference instance and line drawing during the colorization process.}
\label{bg_control2}
\end{figure}

\subsection{Adaptive Decoupled Control Module}
\label{control detail}
In the process of animation coloring, the main focus is usually on the consistency between the foreground and background, as well as the user-controllable friendliness. However, current methods often lack such controllable capabilities and consistency.
To address the critical issue of fine-grained detail degradation in background elements, instance-specific features (\textit{e.g.}, small props, clothing patterns), and text descriptions, we propose the \textit{Adaptive Decoupled Control Module}. 

This module explicitly injects detailed semantic cues into the diffusion process through four coordinated components: information projection modules, expert modules, a learnable dynamic condition weight, and a condition gating layer, ensuring the adaptive preservation of intricate details across diverse scenarios. The core design utilizes CLIP and T5 encoders for robust feature extraction, followed by modality-specific projections and targeted injections via separate cross-attention branches. Its implementation is detailed as follows: First, we encode reference background, instance elements, and text descriptions using modality-specific encoders to capture their distinct semantic and visual details. Let $bg$ denote the reference background, $\{I_1, I_2, ..., I_N\}$ denote $N$ instance elements, and $F_{\text{text}} = \text{T5}(T)$ denote the text description. Their visual raw features are encoded as:
\begin{equation}
    F_{\text{inst}}^i = \text{CLIP}(I_i) \ (\forall i \in \{1, 2, \dots, N\}),
\end{equation}
and $F_{\text{bg}} = \text{CLIP}(bg)$.  
Next, to align these heterogeneous features with the hidden dimension of the Diffusion Transformer backbone, the information projection module processes each modality through a dedicated multi-layer perceptron (MLP) branch to preserve modality-specific details:
\begin{flalign}
    \begin{aligned}
        F_{\text{bg}} &= \text{MLP}_{\text{bg}}(F_{\text{bg}}) \in \mathbb{R}^{d \times D_{\text{DiT}}}, \\
        F_{\text{inst}}^i &= \text{MLP}_{\text{inst}}(F_{\text{inst}}^i) \in \mathbb{R}^{d \times D_{\text{DiT}}} \quad (\forall i \in \{1, 2, ..., N\}), \\
        F_{\text{text}} &= \text{Text\_Projection}(F_{\text{text}}) \in \mathbb{R}^{d_{T5} \times D_{\text{DiT}}},
    \end{aligned}
\end{flalign}
where $d$ is the output dimension of CLIP's vision encoder, and $d_{\text{T5}}$ is the output dimension of T5's encoder. $D_{\text{DiT}}$ is the hidden dimension of DiT's transformer blocks, and $\text{MLP}_{\text{bg}}$, $\text{MLP}_{\text{inst}}$, $\text{Text\_Projection}$ are lightweight, modality-specific MLPs that ensure dimension alignment while retaining fine-grained cues unique to each modality.

For adaptive decoupled control, we design separate cross-attention layers as condition experts for each projected feature, dynamically weighted by a learnable condition weight mechanism. For each DiT transformer block, the hidden states of the sketch sequence $H \in \mathbb{R}^{L \times D_{\text{DiT}}}$ (where $L$ is the sketch patch sequence length) serve as queries, with each modality’s projected feature acting as keys and values in its dedicated cross-attention branch. The aggregated attention output is:
\begin{equation}
    \begin{aligned}
    H_{\text{attn}} = W_{\text{bg}} \cdot \text{CrossAttn}(H, F_{\text{bg}}) \\
                    \quad + \sum_{i=1}^N W_{\text{inst}}^i \cdot \text{CrossAttn}(H, F_{\text{inst}}^i) \\  
                    \quad + W_{\text{text}} \cdot \text{CrossAttn}(H, F_{\text{text}}),
    \end{aligned}
\end{equation}
where $W_{\text{bg}} \in \mathbb{R}^+$ (background weight), $W_{\text{inst}}^i \in \mathbb{R}^+$ (instance-specific weights for each $I_i$), and $W_{\text{text}} \in \mathbb{R}^+$ (text weight) form the learnable weight vector. During training, $W_{\text{bg}}$  is frozen (preserving visual background consistency), while $W_{\text{text}}$ and $W_{\text{inst}}^i$ are learned end-to-end to adapt to scene-specific detail priorities.
Finally, a Condition Gating Layer adaptively adjusts hidden states $H$ and attention hidden states $H_{attn}$, ensuring context-aware detail injection. 

As illustrated in the upper and lower panels of Figure~\ref{fig: control}, our method enables the DiT backbone to explicitly attend to background, instance, and textual details via dedicated decoupled pathways at each diffusion step, thereby empowering users to flexibly modulate the control strength of individual conditioning signals for tailored generation results.
By preserving condition-specific cues and adapting weights to scene complexity, the module retains fine-grained details regardless of instance count or scene diversity, significantly enhancing the fidelity of colorized results. More results of semantic background control are presented in Figure~\ref{semantic_bg}.
\section{Experiment}

\subsection{Implementation Detail}
\label{implementation detail}
\paragraph{Experiment settings.} Our approach is based on the Wan2.1-1.3B and Wan2.1-14B~\citep{wang2025wan} pre-trained weights, trained on the Sakuga42M dataset~\citep{sakuga} and animation film data. The data construction pipeline is detailed in section~\ref{datapipeline}. 
Training utilizes the AdamW optimizer with a learning rate of 2e-5, employing a batch size of 1 on two A800(80GB) GPUs and training with 20000 steps.

\paragraph{Evaluation metrics.} We evaluate performance along multiple dimensions. FID~\citep{unterthiner2018towards} quantifies video quality and natural motion smoothness. CLIP Score~\citep{radford2021learning} measures semantic correspondence between the generated video and the reference video.
For conditional generation tasks involving reference videos, we report Learned Perceptual Image Patch Similarity (LPIPS)~\citep{zhang2018unreasonable} and Structural Similarity Index (SSIM)~\citep{wang2004image} to assess frame-level fidelity. 
LPIPS captures perceptual similarity and sensitivity to blur/texture shifts, while SSIM evaluates the preservation of structural information against ground-truth videos. 
Temporal Consistency (Temporal) evaluates video consistency in the temporal dimension by calculating the average similarity between adjacent frames.

\begin{figure}[htbp]
\centering
\includegraphics[width=\linewidth, keepaspectratio]{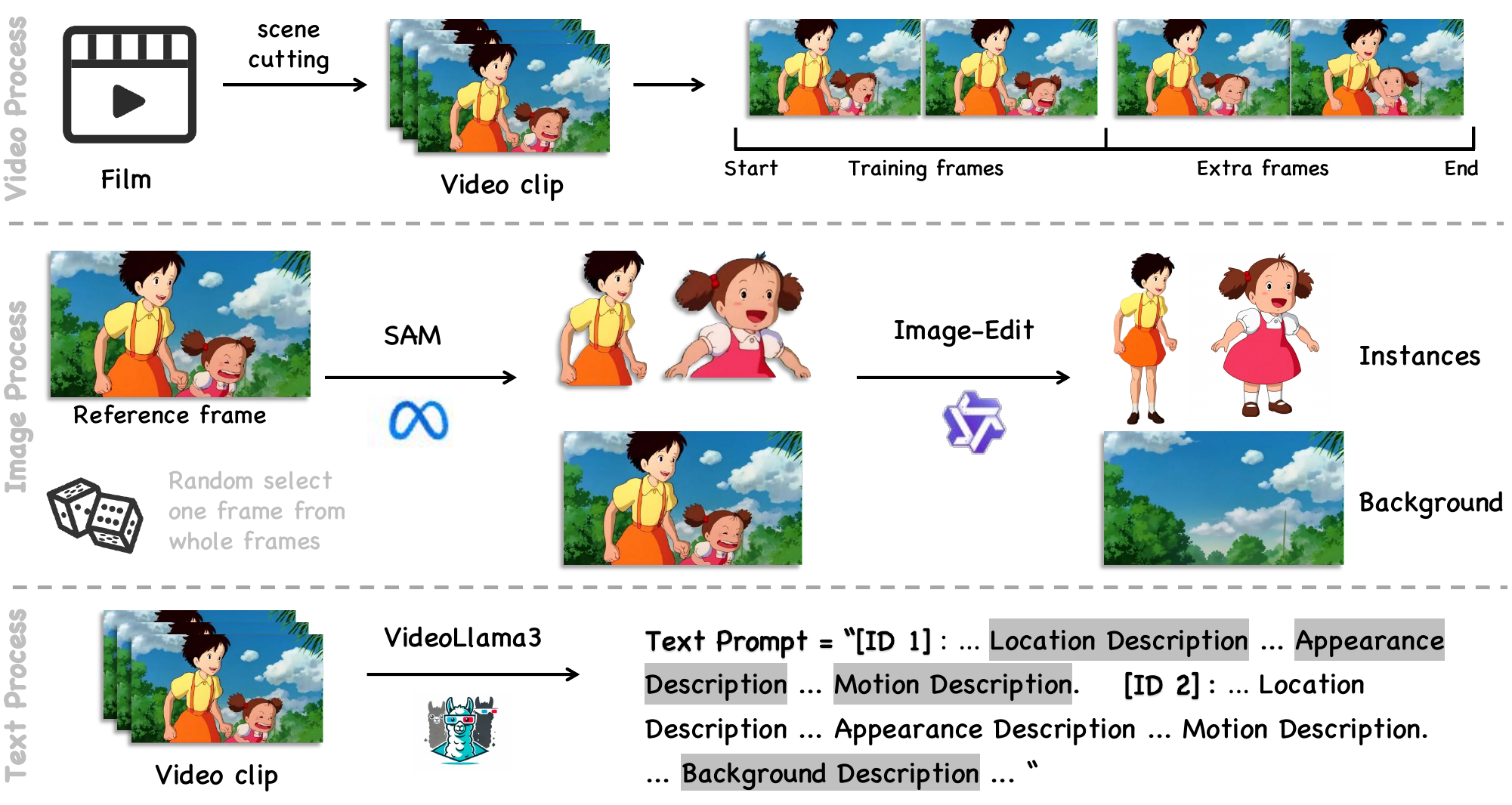}
\caption{\textbf{Data Construction Pipeline}. We build the complete dataset in three steps: Video, Image, and Text Processing.}
\label{data}
\end{figure}

\subsection{Dataset and Pipeline}
\label{datapipeline}
\noindent\textbf{Automated data pipeline construction.}
The data pipeline is designed to automatically process raw data into structured training samples, with three core modules, as follows: 

\noindent\textbf{Video Processing:} Long-form videos (e.g., full-length films, short dramas) first undergo scene segmentation to generate video clips with consistent scene contexts. For each segmented video clip, the first 81 frames are allocated as training data, and the remaining frames are reserved as extra reference frames.

\noindent\textbf{Image Processing:} A single frame is randomly sampled from each video clip (encompassing both training frames and extra reference frames) to serve as the reference frame. The reference frame is processed in three steps: 
\begin{itemize}
    \item \textit{Step 1: Instance Extraction}. Each foreground instance is extracted using the Segment Anything Model (SAM)~\cite{sam}, Recognized Anything Model~\cite{ram}, and GroundingDino~\cite{groundingdino}.
    \item \textit{Step 2: Character Generation}. Qwen-Image-Edit~\cite{qwen-image} is employed to generate complete character images based on the extracted instances.
    \item \textit{Step 3: Background Generation}. Additionally, complete background images are obtained via text-guided editing with Qwen-Image-Edit by removing foreground characters while retaining background details.
\end{itemize}
  
\noindent\textbf{Text Processing:} Finally, the textual description for each video clip is constructed to cover four mandatory dimensions. 
\textit{Location:} spatial position of each character in the video.
\textit{Appearance:} detailed appearance attributes of each character.
\textit{Motion:} actions performed by each character throughout the video. 
\textit{Background:} contextual description of the video background, ensuring comprehensive semantic guidance for model training.

\noindent\textbf{OpenAnimate Dataset}.
The current colorization datasets are generally based on reference frames and contain multiple scenes, which cannot meet the instance-level colorization task. Therefore, we constructed this dataset.
This dataset is designed for the instance-aware sketch video colorization task and aimed at promoting the development of automatic animation techniques. This dataset includes two parts. One is a filter subset from the Sakuga42M dataset~\cite{sakuga}, the other is collected from animation films on the internet. The total video clips are about 42K+, and each video clip is in the same scene, which is cut by scene-cutting algorithms.  For each data clip, we provide a reference frame, multiple reference instances, a background image, and a text description.

\begin{figure*}[h]
    \centering
    \includegraphics[width=\linewidth]{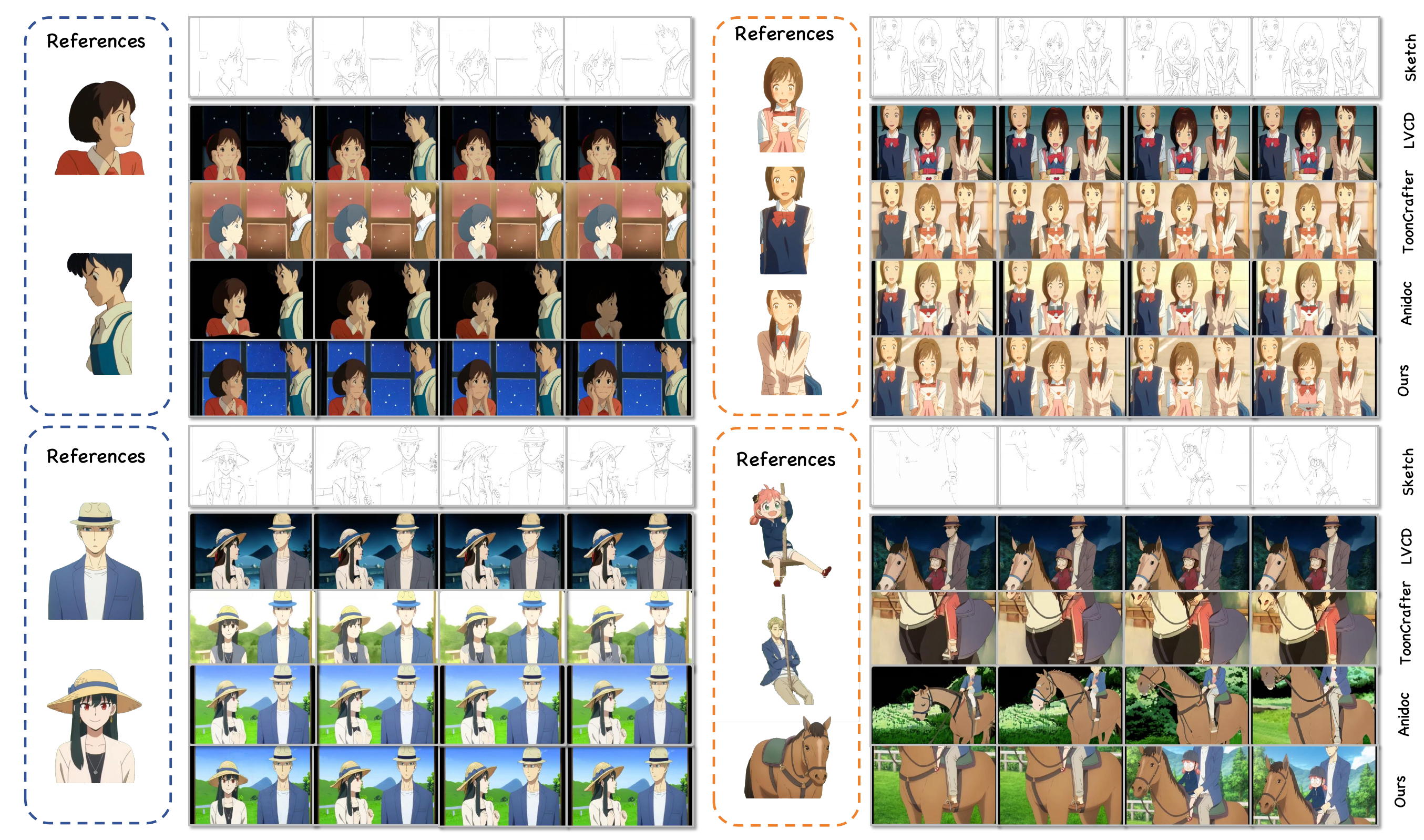}
    \caption{\textbf{Qualitative comparison with baseline methods}. Given diverse reference instances, our method achieves better results in maintaining character controllability and color consistency.}
    \label{fig: compare}
\end{figure*}

\subsection{Quantitative Comparison}
\label{quantitative comparison}
To systematically evaluate the video colorization performance of our approach, we constructed a rigorous test set consisting of 100 high-quality animation video clips with multiple reference instances. This test set is collected from American-style animations, and it was found that the content of all the pictures contains multiple characters after filtering. 
For each clip, we manually extracted character instances as reference design images through SAM, ensuring consistent evaluation conditions across all methods. 

We compare our framework against four recent colorization methods: LVCD~\citep{huang2024lvcd}, ToonCrafter~\citep{xing2024tooncrafter}, Anidoc~\citep{meng2024anidoc}, LayerAnimate~\cite{yang2025layeranimatelayerlevelcontrolanimation}, and ToonComposer~\citep{tooncomposer}. 
To ensure fair comparison while respecting each method's original design, we followed their default input configurations: For LVCD, Anidoc, LayerAnimate, and ToonComposer, we provided the first frame of the original video as the reference image. Two-Stage Animation includes using MangaNinja~\cite{MangaNinja} to colorize the first frame and then using Wan-I2V-14B to colorize the sketches. In contrast, our method uniquely utilizes extracted reference instances for colorizing sketch sequences.
As shown in Table~\ref{tab: comparison}, ToonComposer achieves the best in CLIP Score, while our approach achieves the best performance in the other metrics.
Notably, despite not leveraging the full information of the reference frame, our method provides reference instances and background, which provides more controllability and creative flexibility for users in sketch video colorization.

\begin{table}[t]
    \centering
    \caption{Quantitative comparisons with baselines. \textbf{Bolded}: best, \underline{underscored}: second best.}
    \begin{adjustbox}{width=\columnwidth}
    \begin{tabular}{lcccccc} 
        \toprule
        Model & FID $\downarrow$  & SSIM $\uparrow$ & LPIPS $\downarrow$ & Temporal $\uparrow$ & CLIP $\uparrow$ \\
        \midrule
        ToonCrafter(+First and Last Frame) & 134.497 & \underline{0.661 }& \underline{0.272} & 0.961 & 0.894 \\
        Anidoc(+First Frame) & 305.382 & 0.343 & 0.637 & 0.961 & 0.691 \\
        LVCD(+First Frame) & 138.828 & 0.561 & 0.451 & 0.962 & 0.871 \\
        LayerAnimate(+First Frame) & 184.048 & 0.601 & 0.431 & 0.959 & 0.849 \\
        Two-Stage Animataion(+First Frame) & 245.172 & 0.424 & 0.466 & 0.954 & 0.756 \\
        ToonComposer(+First Frame) & \underline{132.319} & 0.581 & 0.304 & \underline{0.968} & \textbf{0.929} \\
        \textbf{Ours(+Instances)} & \textbf{121.901} & \textbf{0.681} & \textbf{0.201} & \textbf{0.969} & \underline{0.921} \\
        \hline
    \end{tabular}
    \end{adjustbox}
    \label{tab: comparison}
\end{table}

\subsection{Qualitative Comparison}
As depicted in Figure~\ref{fig: compare}, our method demonstrates superiority in generating high-fidelity results. It not only produces significantly clearer textures, such as the fine details of clothing patterns and hair strands, but also better preserves the character’s unique identity, including subtle facial features and postures.

In scenarios where there are substantial differences between the reference instances and input sketches, our method shines. For example, when the input sketch has a large-scale motion transformation compared to the reference, LVCD fails to accurately colorize the sketches, and it cannot follow the large-scale motion in the results, leading to disjointed color mappings. ToonCrafter, on the other hand, struggles when provided with multiple reference instances. It often misinterprets the correct color information, resulting in color conflicts or inappropriate color assignments.
Compared to Anidoc, our method demonstrates advantages in handling complex motion scenarios.
Anidoc shows weakness when the reference instances themselves contain motion and when trying to follow camera trajectories. In such cases, Anidoc's results often lack coherence, with characters or background elements appearing misaligned or discolored.

Furthermore, compared to baseline methods, a key advantage of our approach lies in its superior control over both instances and backgrounds. As illustrated in Figure~\ref{bg_control2}, while keeping reference instances and input sketches fixed, we can freely modify background images, and our colorization results consistently maintain robust consistency between foreground instances and the updated backgrounds. Additionally, our method supports semantic-based background control, where colorization outputs are not constrained to strictly replicate the content of reference images. This flexibility enables broader and more versatile application scenarios. As shown in Figure~\ref{semantic_bg}, we demonstrate semantic-driven background control results under the same background reference image, validating the method’s ability to adapt to textual semantic cues. One point to note is that there is a prerequisite for such a result, which is that the line drawing sequence should not contain background information, and it is best to have a blank background.

\subsection{Ablation Study}
To investigate the contribution of key components in our framework, we conduct systematic ablation experiments by sequentially removing three core modules: Instance Matching, Canvas Guidance, and Adaptive Decoupled Control. 
As shown in Table~\ref{tab: ablation}, the Full Model achieves the best overall performance across most metrics, confirming the effectiveness of our integrated design.
\begin{table}
    \centering
    \caption{Quantitative results of ablation study.}
    \renewcommand{\arraystretch}{1.2}
    \resizebox{0.999\linewidth}{!}{
    \begin{tabular}{lcccccc}
        \hline
        Configuration & FID $\downarrow$ & SSIM $\uparrow$ & LPIPS $\downarrow$ & Temp $\uparrow$ & CLIP $\uparrow$ \\
        \hline
        Full Model & \textbf{121.901} & \textbf{0.681} & \textbf{0.201} & \textbf{0.969} & \textbf{0.921} \\
        w/o Instance Attention & 138.615 & 0.576 & 0.317 & 0.952 & 0.861 \\
        w/o Canvas Guidance & 129.221 & 0.605 & 0.289 & 0.958 & 0.875 \\
        w/o Decoupled Control & 126.984 & 0.591 & 0.293 & 0.961 & 0.862 \\
        \hline
    \end{tabular}}
    \label{tab: ablation}
    \vspace{-5pt}
\end{table}

\begin{figure}
    \centering
    \includegraphics[width=\linewidth]{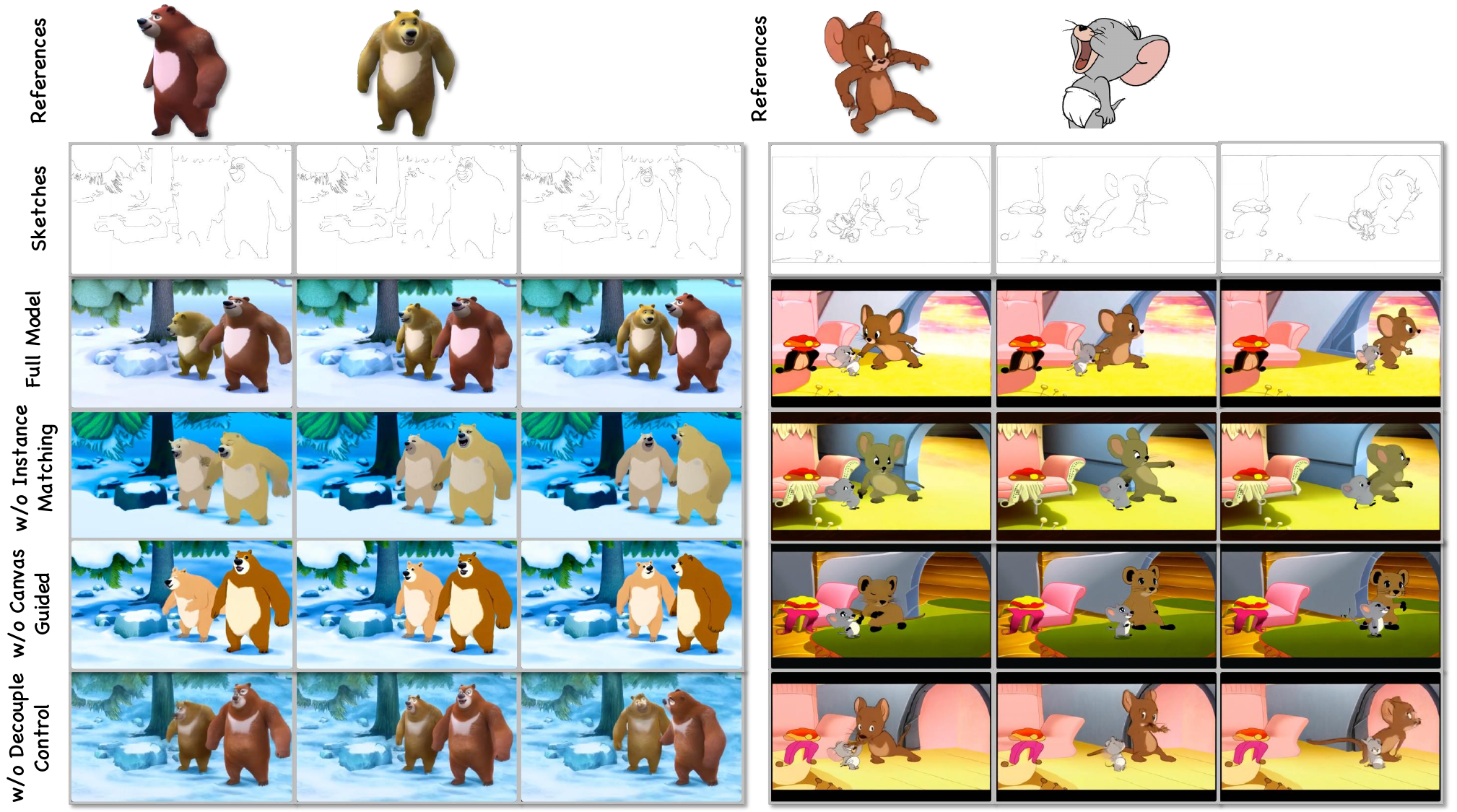}
    \caption{\textbf{Qualitative results of ablation study}. All proposed modules effectively contribute to improving the quality and consistency of multi-instance colorization.}
    \label{fig: ablation}
\end{figure}

\noindent\textbf{Efficacy of Instance-Aware Attention:} As shown in Figure~\ref{fig: ablation}, the degradation is particularly evident in identity preservation results, revealing that characters lose consistent color mapping across frames, with frequent mix-ups between reference instances. This confirms that Instance-Aware Attention is critical for maintaining per-character identity throughout the video. 
As illustrated in Figure~\ref{single_control} and \ref{multi_control}, we demonstrate the character colorization results of our method under reference instances with diverse styles and textures. Specifically, we perform colorization on input sketch videos using references, including various poses of the same character as well as entirely distinct characters. Our outputs closely align with the visual attributes of the reference inputs. For multi-person scenarios, we select reference instances with drastically different styles and identities for sketch video colorization, and our method still maintains robust character consistency. Notably, the result is not influenced by shuffle or reorder instances.

\noindent\textbf{Efficacy of Canvas Guidance:} Removing this module results in higher FID and lower SSIM compared to the Full Model, which means that the user canvas provides useful information in generation. Visually, this leads to noticeable color inconsistency, especially in background regions and instances. On this basis, we also observed that there are significant differences in the animation styles of the characters.

\noindent\textbf{Efficacy of Adaptive Decoupled Control:} This ablation causes more modest performance drops but introduces distinct detail degradation issues. As shown in Figure~\ref{fig: ablation}, fine-grained details such as facial expressions, fabric textures, and accessory patterns become blurred or inconsistent. With the Adaptive Decoupled Control Module, the foreground and background have better integration.

\section{User Study}
\label{user study}
To evaluate the practical performance of our method, we designed a user study comparing it with three representative baseline methods across four core user-centric dimensions. 20 participants (aged 18–60, 60\% of whom are female) were recruited, all of whom had adequate knowledge and experience in computer vision, line art drawing, or related fields. This ensures participants have sufficient expertise to distinguish subtle differences in colorization quality and tool usability. All participants provided informed consent and were compensated according to local standards.

\begin{figure}[htbp]
\centering
\includegraphics[width=\linewidth, keepaspectratio]{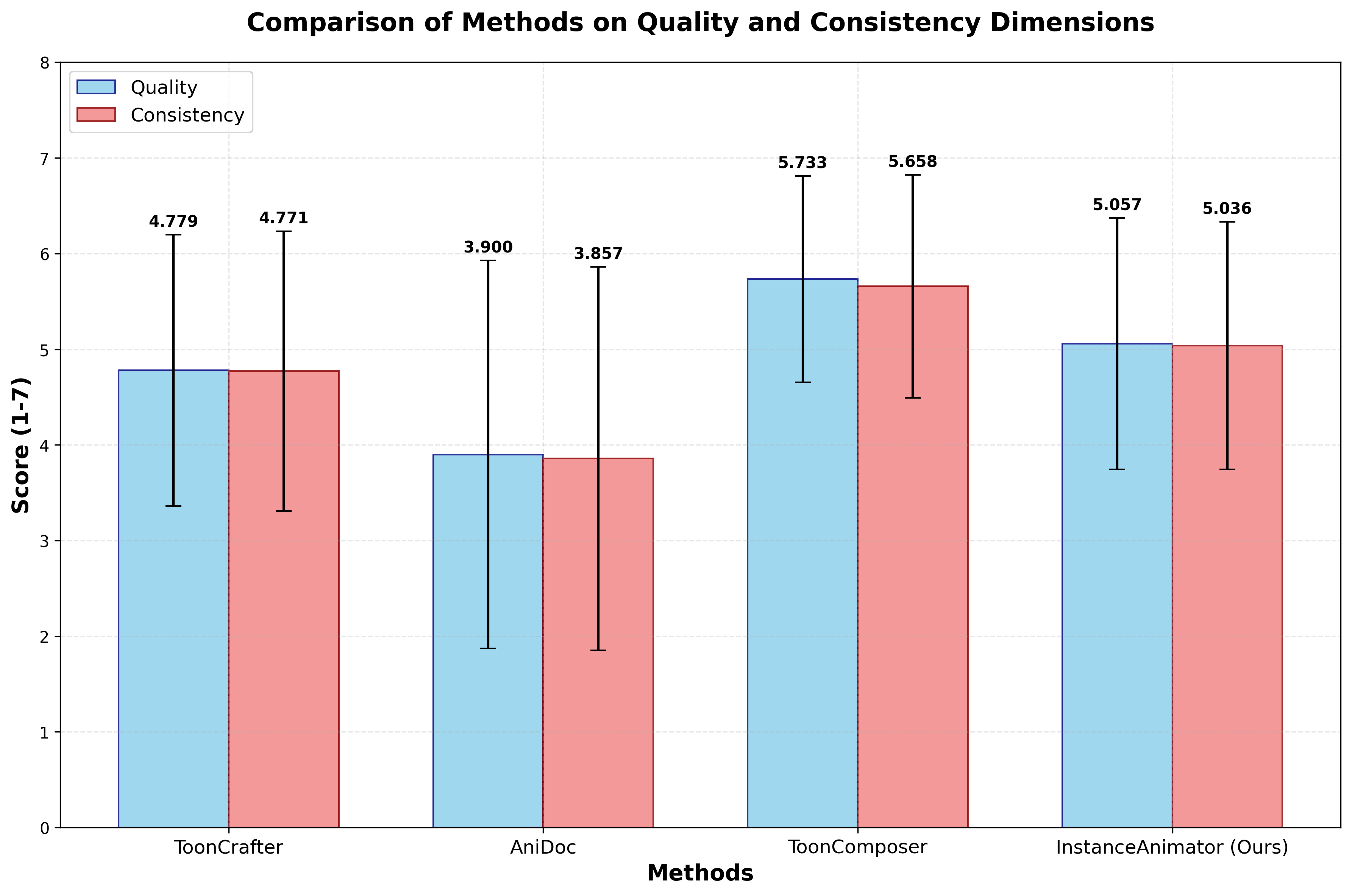}
\caption{\textbf{User ratings across four key dimensions: Our method vs. three baselines}. 
All ratings are based on a 7-point Likert scale (1 = Strongly Disagree, 7 = Strongly Agree).}
\label{user_study}
\end{figure}

The study was conducted remotely by observing the results of coloring the same set of line drawings using different methods and experiencing different colorization workflows, which lasted 20 minutes per participant. We applied the generated dataset for a comparative study in the Main Sections~\ref{implementation detail}-\ref{quantitative comparison}. We randomly sampled 27 video clips from the 100 pairs of examples for each participant. Participants were presented with one video at a time (randomly shuffled) and asked to assess the quality and consistency of each video clip on 7-point scales. After this, participants experienced two interfaces of reference-based (AniDoc) and instance-based (ours) colorization workflows in random order and evaluated their usability and controllability on 7-point scales.

Based on user needs and perception goals, the four evaluation dimensions are defined as follows:
\begin{itemize}
\item \textbf{Quality}: Measures the visual naturalness, accuracy, and overall satisfaction of the colorized output. Participants rated their agreement with the statement: “I am satisfied with the final colorization effect.”
\item \textbf{Consistency}: Assesses color stability across consecutive video frames. Participants rated agreement with: “I think the colors between video frames are very consistent.”
\item \textbf{Usability}: Evaluates the ease of operating the workflow. Participants rated agreement with: “This workflow is easy to learn and use.”
\item \textbf{Controllability}: Quantifies the user’s ability to adjust and refine colorization results. Participants rated agreement with: “I feel I can control the colorization results well.”
\end{itemize}

We collected the results and calculated the average ratings for each method by dimension. As shown in Figure~\ref{user_study}, while our \textit{InstanceAnimator} ranks second only to ToonComposer in quality and consistency, our method has significantly greater efficiency, using 14 times fewer parameters and without requiring the whole first frame as a condition that ToonComposer relies upon. In terms of usability and controllability, our instance-based approach is also ahead of the reference-based approach (5.350±1.268 vs. 5.250±1.832 for usability; 5.300±1.302 vs. 5.000±1.974 for controllability) because we can achieve disentangled instance and background controls that are more in line with practical applications. Participants also conducted creative experimentation using both interfaces. Based on their feedback, the highest-rated features of our \textit{InstanceAnimator} are the support for intuitive animation workflow, achieving longer temporal coherence with minimal reference requirements.

\begin{table*}[h]
    \centering
    \small
    \setlength{\tabcolsep}{4pt}
    \caption{\textbf{Functional Comparison.} Detailed comparison of different video colorization approaches across key functional dimensions. $\checkmark$: full support, $\triangle$: partial/limited support, \ding{55}: not supported.}
    \label{tab:functional_comparison}
    \begin{tabular}{lccccc}
    \toprule
    \textbf{Capability} & \textbf{LVCD} & \textbf{ToonCrafter} & \textbf{Anidoc} & \textbf{ToonComposer} & \textbf{Ours} \\
    \midrule
    \textbf{Reference Type} & Single frame & First+last frame & Single frame & Single frame & Instance-based \\
    \textbf{Multi-Character} & $\triangle$ & $\triangle$ & $\triangle$ & $\checkmark$ & $\checkmark$ \\
    \textbf{Canvas Placement} & \ding{55} & \ding{55} & \ding{55} & \ding{55} & $\checkmark$ \\
    \textbf{Large Motion} & \ding{55} & $\triangle$ & \ding{55} & $\triangle$ & $\checkmark$ \\
    \textbf{Element Editing} & \ding{55} & \ding{55} & \ding{55} & \ding{55} & $\checkmark$ \\
    \textbf{Text Control} & \ding{55} & $\checkmark$ & $\checkmark$ & $\checkmark$ & $\checkmark$ \\
    \textbf{Temporal Consistency} & $\triangle$ & $\checkmark$ & $\triangle$ & $\checkmark$ & $\checkmark$ \\
    \textbf{Detail Quality} & $\triangle$ & $\triangle$ & $\triangle$ & $\checkmark$ & $\checkmark$ \\
    \bottomrule
    \end{tabular}
\end{table*}

\section{Functional Comparison}
\label{funtional comparison}

Table~\ref{tab:functional_comparison} presents a comprehensive comparison of our \textit{InstanceAnimator} against existing video colorization methods across eight key functional dimensions. The comparison reveals significant limitations in current approaches: traditional methods like LVCD~\citep{huang2024lvcd} and Anidoc~\citep{meng2024anidoc} rely on single-frame references and struggle with multi-character scenarios, large motion handling, and advanced editing capabilities. While ToonCrafter~\citep{xing2024tooncrafter} introduces first-and-last frame guidance and ToonComposer~\citep{tooncomposer} achieves better multi-character support, both methods still lack canvas placement control and element-level editing functionality. In contrast, our \textit{InstanceAnimator} demonstrates superior capabilities across all dimensions through its novel instance-based reference system. Notably, we are the only method that supports canvas placement and element editing, enabling precise control over character positioning and fine-grained modifications. The instance-based approach also ensures robust handling of large motions and maintains high temporal consistency while preserving detail quality, making \textit{InstanceAnimator} a comprehensive solution for professional video colorization workflows.

\section{Conclusion}
In this paper, we propose \textit{InstanceAnimator}, a novel DiT-based framework for multiple reference instances sketch video colorization. Our method includes Canvas Guidance Condition and Instance Matching Mechanism to establish a correspondence between line art sequences and reference instances and unify reference instances on a blank canvas as a spatio-temporal anchor to resolve temporal misalignment issues. Meanwhile, the proposed Adaptive Decoupled Control Module injects fine-grained background and instance details into the diffusion process, preserving intricate features like textures and patterns. We demonstrate through extensive experiments that our approach supports diverse instances and improves model generalization in the colorization process.

{
    \small
    \bibliographystyle{ieeenat_fullname}
    \bibliography{main}
}

\clearpage
\appendix
\section*{Appendix}

\section{Instance Control}
\label{instance control}
Instance control is a core technical bottleneck in animation video colorization, as it directly determines the accuracy of color transfer between reference instances and target sketch characters, and further affects the consistency and naturalness of the final colorization results. From the perspective of identity correlation and visual consistency between reference and target, instance control in this task can be systematically categorized into two typical scenarios: (i) Cross-identity control, where the reference instance and the target sketch character belong to different identities, leading to significant discrepancies in facial features, contour structures, and overall appearance, which pose great challenges to accurate feature alignment and reliable color transfer; (ii) Intra-identity control, where the reference instance and the target sketch character share the same identity but differ in pose, gesture, or attire, and the inherent visual similarity between them enables relatively stable and accurate colorization results through appropriate alignment mechanisms.
Current research on instance control for animation video colorization mainly revolves around two technical paradigms, \textit{Point-Matching-Based} and the \textit{Text-ID Alignment} framework, which are widely adopted in colorization tasks.

\paragraph{Point-Matching-Based Paradigm.} A dominant technical route in existing studies relies on precise point-matching-based alignment mechanisms, which achieve color propagation by establishing pixel-level or keypoint-level positional correspondence between the reference image (typically a color first frame) and subsequent sketch frames. This paradigm is inherently constrained to single-reference scenarios (e.g., MangaNinja~\cite{MangaNinja}, LayerAnimate~\cite{yang2025layeranimatelayerlevelcontrolanimation}), as its effectiveness heavily depends on the strong positional dependencies provided by a single fixed reference frame. Although some works have attempted to extend this idea to multi-reference scenarios by introducing decoupled control under the U-Net architecture, the integration of multi-reference instance control into the DiT framework— which is more suitable for video sequence modeling—remains a largely underexplored research gap, limiting the scalability of such methods in complex animation colorization tasks.

\paragraph{Text-ID Alignment Paradigm.} Another emerging technical paradigm draws on advances in ID customized video generation, leveraging text-ID alignment and fusion to enhance instance control capability. This line of research focuses on identity preservation, which is also the core inspiration for our proposed method. Recent representative works~\cite{zhang2025easycontrol,zhang2024ssr, z1, z2, z3, z17, z18,  chen2025s2guidancestochasticselfguidance, chen2025taming, song2025makeanything},  include ConsistID~\cite{consisid}, which adopts Q-Former to fuse facial features and keypoint features for strengthening identity representation, and CineMaster~\cite{wang2025cinemaster}, which integrates class labels, 3D bounding boxes, and detailed textual descriptions, and leverages ControlNet to reinforce identity control. However, these methods suffer from inherent technical limitations: Q-Former relies on fixed-dimensional learnable queries, requiring large-scale annotated training data to ensure effective feature learning, while its expressive capacity is constrained by the fixed query dimension; ControlNet, on the other hand, replicates entire DiT blocks to achieve control, resulting in substantial training parameter overhead and reduced computational efficiency, making it difficult to deploy in lightweight or real-time colorization scenarios.

To address the aforementioned limitations of existing methods, including the single-reference constraint of point-matching-based approaches, the high data dependency and computational overhead of text-ID fusion methods, and the lack of effective multi-reference integration under the DiT framework. We propose an Instance-Aware Attention and Adaptive Decouple Control Module, which establishes a more concise, efficient alignment paradigm that is fundamentally different from traditional alignment or point-tracking-based colorization strategies.

The instance control capability of our method is anchored in two complementary core mechanisms, which jointly ensure accurate cross-modal alignment and reliable instance-specific color transfer:
(1) Visual Reference: This mechanism focuses on learning the intrinsic correlation between reference instances (single or multiple) and input sketch frames, establishing a direct visual correspondence that is not limited by strict positional constraints. By capturing the global and local visual features of reference instances, it effectively mitigates the impact of pose or attire variations in intra-identity scenarios and reduces misalignment in cross-identity scenarios.
(2) Textual Guidance: Seamlessly integrated into the data pipeline design, this mechanism provides precise semantic constraints for instance alignment. Specifically, during caption construction, we explicitly describe each instance’s spatial location, detailed appearance attributes (e.g., color, texture, style), and identity-related information in natural language, converting high-level semantic requirements into interpretable feature cues that complement visual information.

\begin{figure*}[htbp]
\centering
\includegraphics[width=\linewidth, keepaspectratio]{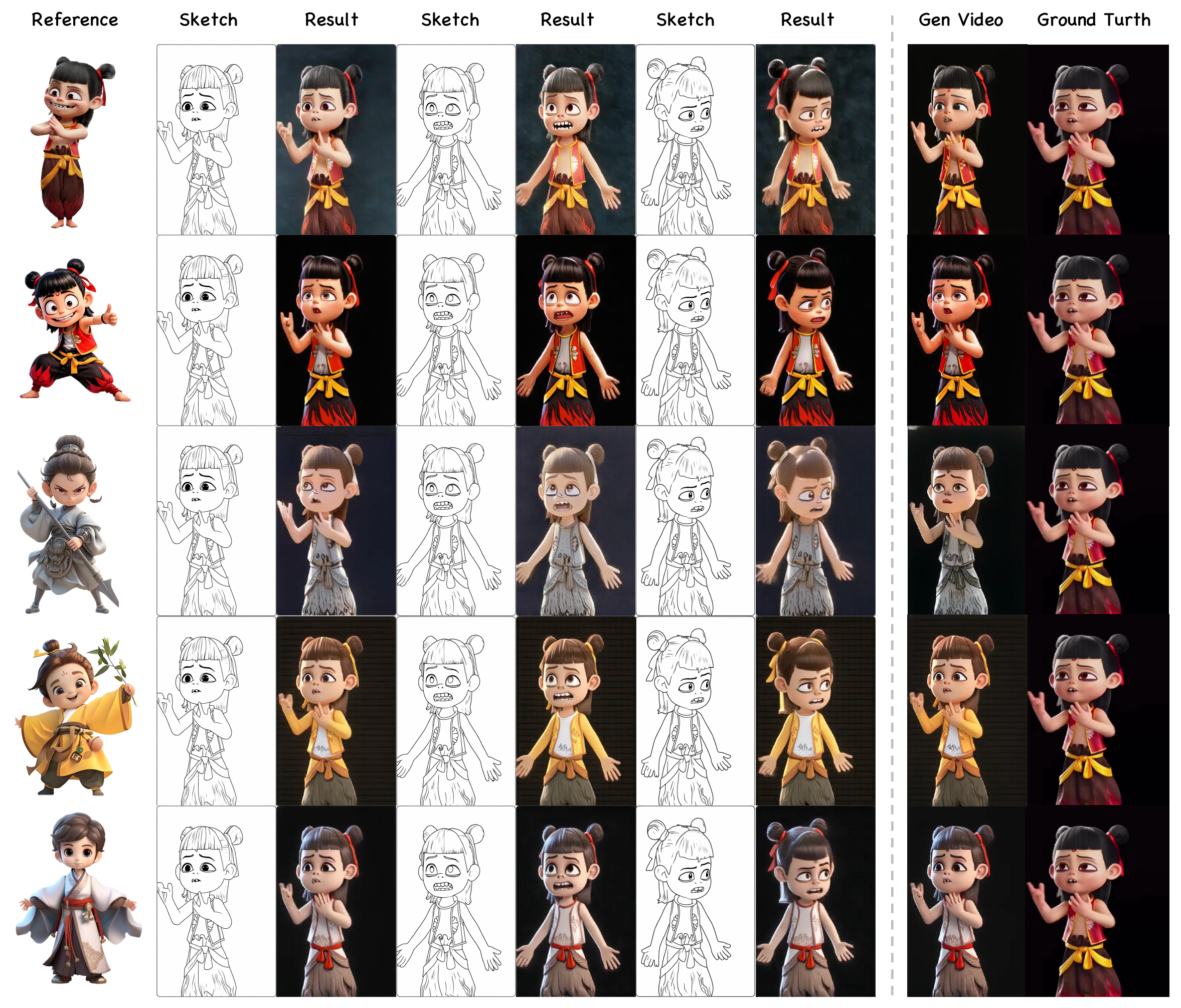}
\caption{\textbf{Colorization from Different Style Reference}.  For the same character, our method easily learns features. For different style characters, our method can transfer their style to the sketches even without prior knowledge.}
\label{single_control}
\end{figure*}

\begin{figure*}[htbp]
\centering
\includegraphics[width=\linewidth, keepaspectratio]{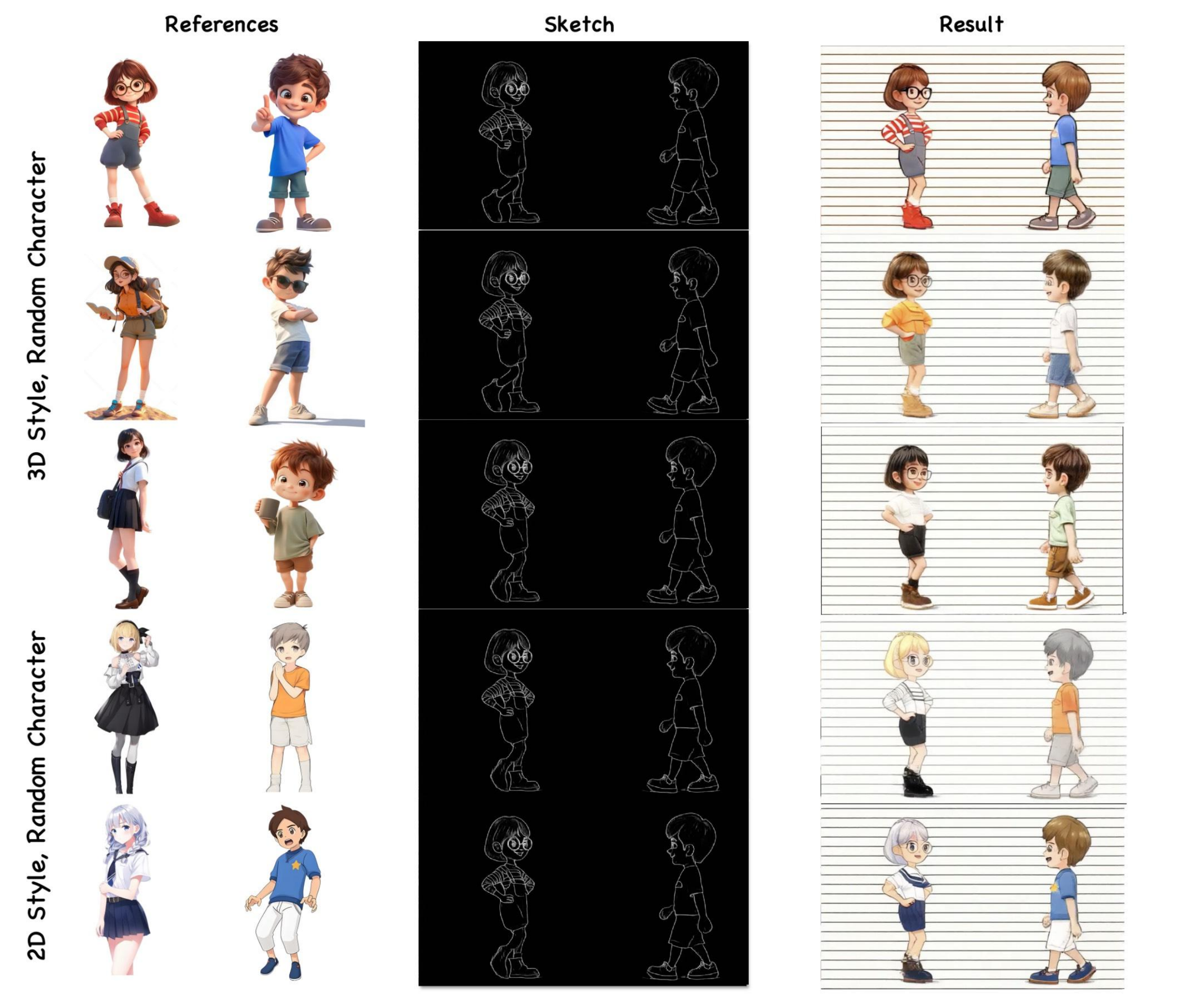}
\caption{\textbf{Multiple Reference Instances Control}. While we obtain different style reference instances from the internet, our method can match the most similar ID and follow the reference instance style to colorize the sketch sequence.}
\label{multi_control}
\end{figure*}

As shown in Fig.~\ref{single_control} and \ref{multi_control}, we present diverse reference instances across distinct styles to validate our model’s instance control capability. Specifically, in Fig.~\ref{single_control} (single-instance control scenario), we vary the reference instances from the same identity to completely distinct identities. Our generated results consistently adapt to changes in the visual references, reflecting precise instance alignment. For Fig.~\ref{multi_control} (multi-instance control scenario), we extend the setting to multiple reference instances whose identities and styles are fully inconsistent with those of the input sketches. Even under this challenging condition, our results still demonstrate strong generalization by transferring the style of the references to the colorization output. Notably, when identities and styles are completely mismatched, our model primarily leverages the overall style of the references rather than strict pixel-wise color alignment. This is because one-to-one visual correspondence between references and sketches is absent, making precise color matching infeasible.

\begin{figure*}[htbp]
\centering
\includegraphics[width=\linewidth, keepaspectratio]{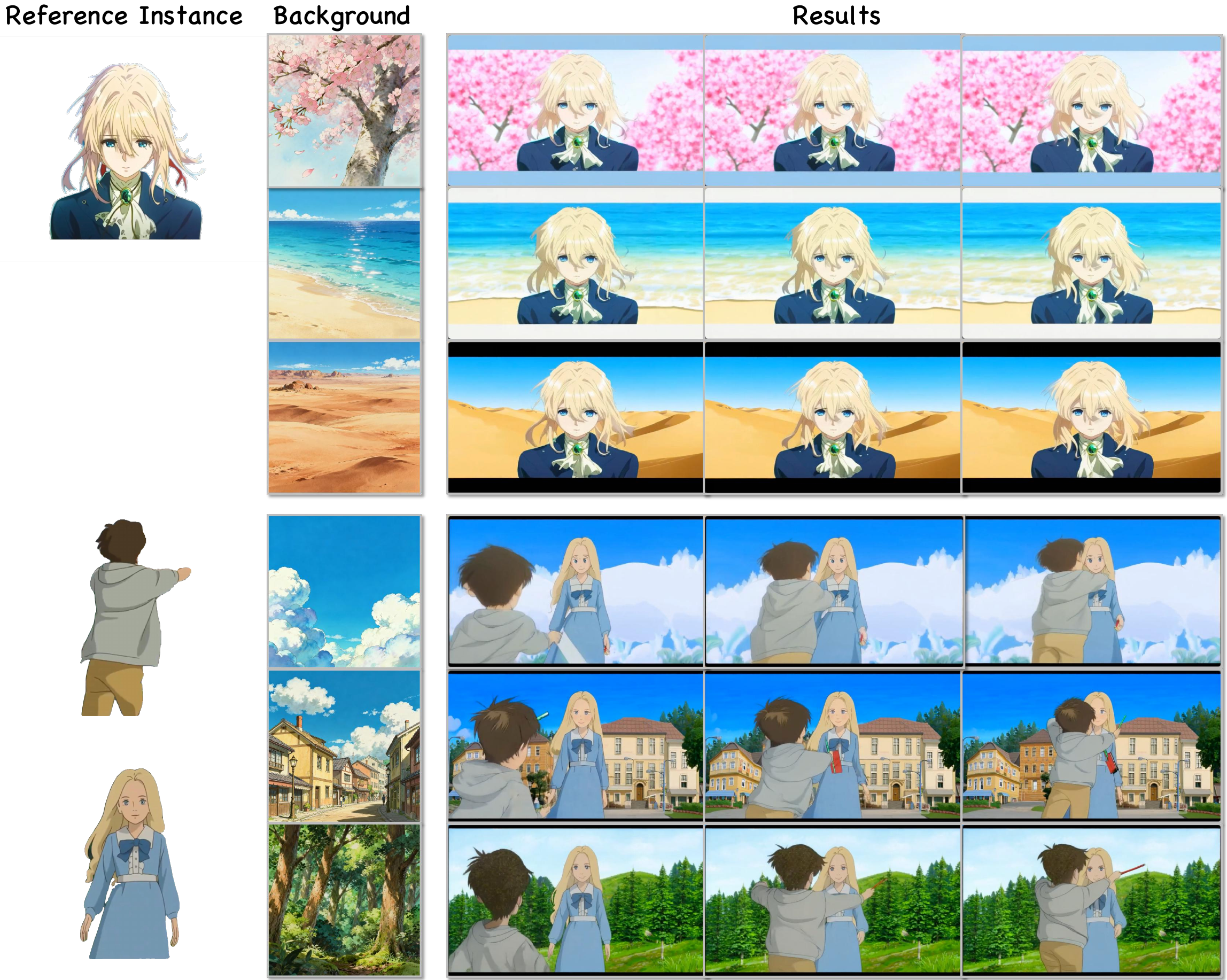}
\caption{\textbf{Semantic background control}. Unlike traditional methods that rely on other tools for directly replacing background, \textit{InstanceAnimator} supports semantic mapping of background in combination with line drawing during the colorization process.}
\label{semantic_bg}
\end{figure*}

\section{Background Control}
\label{sec: bg control}

Effective background colorization is a pivotal yet frequently neglected aspect of video line art colorization. Conventional methods typically offer no native support for background customization; users must rely on external tools to manually replace backgrounds, a fragmented workflow that often introduces visual discontinuities in style, lighting, or semantic context between foreground subjects and the inserted background. This limitation severely constrains practical usability and stifles creative flexibility.

\textit{InstanceAnimator} resolves this challenge by integrating background control directly into the colorization pipeline through two complementary mechanisms. First, \textbf{Visual Background Control} (Figure~\ref{bg_control2}) allows users to provide a custom background image, which is harmoniously fused with the reference instance and input line drawing to generate a visually coherent output. Second, \textbf{Semantic Background Control} (Figure~\ref{semantic_bg}) enables context-aware background modification via textual prompts or semantic mapping (e.g., transforming "indoor studio" to "sunset garden") while preserving structural alignment with the line art, eliminating the need for pixel-level replacement. Unlike traditional approaches that treat background editing as a disjointed post-processing step, our method performs background adaptation intrinsically during colorization, ensuring holistic scene consistency and eliminating artifacts from manual compositing. This unified design empowers users with unprecedented creative agency: backgrounds can be tailored to narrative needs, stylistic preferences, or environmental context without compromising temporal smoothness or artistic integrity. As validated in our experiments, this capability significantly enhances the method's applicability in professional animation workflows, interactive content creation, and personalized media production.

\end{document}